\documentclass[11pt,a4paper]{article}

\usepackage[utf8]{inputenc}
\usepackage[T1]{fontenc}
\usepackage{amsmath,amssymb}
\usepackage{booktabs}
\usepackage{graphicx}
\usepackage{xcolor}
\usepackage{hyperref}
\usepackage{cleveref}
\usepackage{multirow}
\usepackage{tabularx}
\usepackage[margin=2.5cm]{geometry}
\usepackage{enumitem}
\usepackage{caption}
\usepackage{subcaption}
\usepackage{authblk}

\hypersetup{colorlinks=true, linkcolor=blue!60!black, citecolor=blue!60!black, urlcolor=blue!60!black}

\newcommand{\modelname}[1]{\textsc{#1}}
\newcommand{\metricbf}[1]{\textbf{#1}}

\title{Multilingual Multi-Label Emotion Classification\\at Scale with Synthetic Data}

\author{Vadim Borisov\thanks{Email: \href{mailto:vadim@tabularis.ai}{vadim@tabularis.ai}}}
\affil{\href{https://tabularis.ai}{tabularis.ai}}
\date{}

\begin{document}
\maketitle

\begin{abstract}
Emotion classification in multilingual settings remains constrained by the  scarcity of annotated data: existing corpora are predominantly English,  single-label, and cover few languages. We address this gap by constructing  a large-scale synthetic training corpus of over 1M multi-label samples  (50k per language) across 23~languages: Arabic, Bengali, Dutch, English, 
French, German, Hindi, Indonesian, Italian, Japanese, Korean, Mandarin,  Polish, Portuguese, Punjabi, Russian, Spanish, Swahili, Tamil, Turkish,  Ukrainian, Urdu, and Vietnamese, covering 11~emotion categories using culturally-adapted generation and programmatic quality filtering. We train and compare six multilingual transformer encoders, from \modelname{DistilBERT} (135M parameters) to \modelname{XLM-R-Large} (560M parameters), under identical conditions. On our in-domain test set, \modelname{XLM-R-Large} achieves 0.868~F1-micro and 0.987~AUC-micro. To validate against human-annotated data, we evaluate all models zero-shot on GoEmotions (English) and SemEval-2018 Task~1 E-c (English, Arabic, Spanish). On threshold-free ranking metrics, \modelname{XLM-R-Large} matches or exceeds English-only specialist models tying on AP-micro (0.636) and LRAP (0.804) while surpassing on AUC-micro (0.810 vs.\ 0.787) while natively supporting all 23~languages. The best base-sized model is publicly available.\footnote{\url{https://huggingface.co/tabularisai/multilingual-emotion-classification}}
\end{abstract}

\section{Introduction}\label{sec:intro}

Emotion classification is among the most studied tasks in understanding natural languages, with direct applications in mental health monitoring, customer feedback analysis, and conversational artificial intelligence (AI)~\cite{li2014text, plaza2024emotion}. However, the vast majority of research targets English~\cite{demszky2020goemotions,mohammad2018semeval}, and existing multilingual resources when they exist at all, typically cover only a handful of languages with single-label annotation. This leaves a significant gap for real-world systems that must operate across languages and handle the inherent co-occurrence of emotions in natural text.

Building multilingual, multi-label emotion classifiers faces two core challenges:

\textbf{Annotation scarcity.} Labelling emotion data requires native-speaker fluency and cultural competence. Scaling to dozens of typologically diverse languages is prohibitively expensive, which is why most corpora remain English-only~\cite{demszky2020goemotions} or cover at most three languages~\cite{mohammad2018semeval}.

\textbf{Cultural variation.} Emotions are not expressed uniformly across languages. Idiomatic phrases, social display norms, and pragmatic conventions differ substantially, a naïve translation-based approach to corpus construction fails to capture these differences \cite{jackson2019emotion}.

We address both challenges by constructing a large-scale synthetic corpus using culturally-adapted generation. For each of 23~languages spanning 9~language families and 9~scripts, we generate 50k~multi-label examples with language-specific prompting and programmatic quality filtering, yielding a training set of over 1M~samples. We train six multilingual transformer encoders under identical conditions and evaluate them zero-shot on GoEmotions~\cite{demszky2020goemotions} and SemEval-2018 E-c~\cite{mohammad2018semeval}. Our contributions are:

\begin{itemize}[nosep]
    \item \textbf{Large-scale corpus.} A culturally-adapted synthetic dataset of over 1M multi-label samples across 23~languages and 11~emotion classes.
    \item \textbf{Encoder comparison.} A controlled comparison of six multilingual transformers establishing compute--quality trade-offs.
    \item \textbf{Cross-benchmark transfer.} Models trained entirely on synthetic data match English-only specialists on threshold-free ranking metrics---while natively supporting 23~languages.
    \item \textbf{Public release.} The best base-sized model is publicly available.
\end{itemize}
\section{Dataset Construction}\label{sec:dataset}

This section describes the construction of our synthetic multilingual emotion corpus. The pipeline consists of four stages: (1)~definition of the emotion taxonomy, (2)~language selection, (3)~culturally-adapted generation with quality filtering, and (4)~corpus assembly and splitting. We describe each component below.

\subsection{Emotion Taxonomy}

We define 11~emotion categories: \emph{anger}, \emph{contempt}, \emph{disgust}, \emph{fear}, \emph{frustration}, \emph{gratitude}, \emph{joy}, \emph{love}, \emph{neutral}, \emph{sadness}, and \emph{surprise}. This taxonomy extends Ekman's six basic emotions~\cite{ekman1992basic} with finer-grained social emotions (\emph{contempt}, \emph{frustration}, \emph{gratitude}, \emph{love}) and a \emph{neutral} class. Compared to GoEmotions~\cite{demszky2020goemotions}, which defines 28~fine-grained English-only classes, and SemEval-2018 E-c~\cite{mohammad2018semeval}, which uses 11~classes across 3~languages, our taxonomy is designed to balance granularity with cross-lingual applicability. Importantly, we adopt a multi-label formulation: each sample may carry one or more emotion labels simultaneously, reflecting the well-documented co-occurrence of emotions in natural text~\cite{mohammad2018semeval}.

\subsection{Language Selection}

The corpus covers 23~languages: Arabic, Bengali, Dutch, English, French, German, Hindi, Indonesian, Italian, Japanese, Korean, Mandarin, Polish, Portuguese, Punjabi, Russian, Spanish, Swahili, Tamil, Turkish, Ukrainian, Urdu, and Vietnamese. These were chosen to maximise typological and geographic diversity, spanning 9~language families, 8~scripts, and all inhabited continents. The selection includes high-resource languages (English, Mandarin), medium-resource languages (Turkish, Vietnamese), and lower-resource languages (Swahili, Punjabi) to ensure broad coverage.

\subsection{Culturally-Adapted Generation}

For each language, we generate 50k~samples using a pipeline designed to produce culturally authentic emotional text rather than translated English.

\textbf{Cultural tailoring.} Generation prompts are constructed per language to elicit culturally appropriate scenarios, social contexts, and idiomatic expressions. For instance, prompts for Japanese encourage references to hierarchical social obligations, while prompts for Spanish may elicit familial contexts. Emoji usage is included where culturally appropriate but is not a primary focus of the generation strategy.

\textbf{Script diversity.} For languages where Latin-script writing is widespread in informal digital communication notably Hindi, Bengali, Tamil, Punjabi, Urdu, and Arabic---at least 10\% of the generated samples are provided in romanised form. This reflects the common real-world practice of writing these languages in Latin characters on social media and in messaging apps~\cite{gharami2025modeling}.

\textbf{Multi-label construction.} Each sample is annotated with one or more emotion labels. The resulting label cardinality is 1.65~labels per sample on average: approximately 50\% of samples carry a single label, 35\% carry two labels, and 15\% carry three. No empty-label rows exist in the corpus.

\paragraph{Quality filtering.} We apply programmatic checks to remove low-quality or mislabelled samples from our synthetic datasets following best practices. These include lexical diversity thresholds to discard near-duplicate or degenerate outputs, label text consistency verification using auxiliary classifiers, and multi-prompt generation strategies to increase stylistic and topical variety across the corpus.

\subsection{Corpus Statistics}

The final corpus contains over 1.15M~training samples (50k per language), with 500~validation and 500~test rows per language (11,500~each). Splits are stratified by language. The median text length is 209~characters (p95~$\approx$~650~characters); we truncate inputs to 192~tokens during training. \Cref{tab:class_dist} shows the class distribution. Frequencies range from \emph{sadness} (19.0\%) to \emph{neutral} (8.0\%), an imbalance ratio of approximately $2.4\times$---mild enough to not require class re-weighting or oversampling.

\begin{table}[t]
\centering
\caption{Emotion class distribution in the training set.}\label{tab:class_dist}
\small
\begin{tabular}{lrr}
\toprule
\textbf{Emotion} & \textbf{Count} & \textbf{Share (\%)} \\
\midrule
sadness     & 111,059 & 19.0 \\
anger       & 107,058 & 18.3 \\
frustration & 104,456 & 17.9 \\
surprise    &  99,942 & 17.1 \\
disgust     &  92,825 & 15.9 \\
love        &  84,870 & 14.5 \\
fear        &  80,464 & 13.8 \\
contempt    &  80,447 & 13.8 \\
gratitude   &  78,812 & 13.5 \\
joy         &  78,481 & 13.4 \\
neutral     &  46,728 &  8.0 \\
\bottomrule
\end{tabular}
\end{table}
\section{Experimental Setup}\label{sec:setup}

\subsection{Models}

We train six configurations of multilingual transformer encoders, summarised in \Cref{tab:models}.

\begin{table}[t]
\centering
\caption{Models trained. All share the same head, loss, and hyperparameters.}\label{tab:models}
\small
\begin{tabular}{lrrr}
\toprule
\textbf{Model} & \textbf{Params} & \textbf{Epochs} & \textbf{Train (min)} \\
\midrule
\modelname{DistilBERT-multi} \cite{Sanh2019DistilBERTAD}     & 135M & 3 &  14.0 \\
\modelname{mBERT \cite{devlin2019bert}}                & 178M & 3 &  27.4 \\
\modelname{XLM-R-Base \cite{conneau2020xlmr}}           & 278M & 3 &  31.5 \\
\modelname{Twitter-XLM-R} \cite{barbieri-etal-2022-xlm}        & 278M & 3 &  69.8 \\
\modelname{mDeBERTa-v3} \cite{he2021debertav3}          & 278M & 3 &  69.9 \\
\modelname{XLM-R-Large \cite{conneau2020xlmr}}          & 560M & 3 & 130.8 \\
\bottomrule
\end{tabular}
\end{table}

\subsection{Training Details}

All models are trained with binary cross-entropy loss (\texttt{BCEWithLogitsLoss}) using AdamW optimisation \cite{loshchilov2017decoupled}, learning rate $2 \times 10^{-5}$, cosine schedule with 6\% warmup, weight decay 0.01, and gradient clipping at 1.0.Training uses bf16 mixed precision on a single NVIDIA A100 80\,GB GPU with batch size 128 (256 for evaluation). The decision threshold is $\tau = 0.5$; per-model calibration on the validation set yields optimal thresholds between 0.45--0.50, with negligible F1 improvement ($\leq$0.1~points).

\subsection{Evaluation Metrics}

Multi-label classification requires metrics beyond standard accuracy.
We report:
\begin{itemize}[nosep]
    \item \textbf{Threshold-based} ($\tau = 0.5$): subset accuracy (exact match), Hamming accuracy, Jaccard similarity (sample-averaged), F1-micro, and F1-macro.
    \item \textbf{Threshold-free}: micro-averaged AUROC, micro-averaged average precision (AP), and label ranking average precision (LRAP).
\end{itemize}
Threshold-free metrics are particularly important for cross-corpus comparison, where different decision rules (sigmoid $\geq \tau$ vs.\ softmax argmax) make threshold-based F1 non-comparable across model types (\Cref{sec:headtohead}).

\subsection{Cross-Benchmark Evaluation Protocol}\label{sec:protocol}

To validate against human-annotated data, we evaluate all models zero-shot on GoEmotions~\cite{demszky2020goemotions} and SemEval-2018 Task~1 E-c~\cite{mohammad2018semeval} (English, Arabic, Spanish).
Since each benchmark uses a different label inventory, we report results in two label spaces:
\begin{itemize}[nosep]
    \item \textbf{Projected}: collapse the benchmark's native labels into our 11 via a many-to-one mapping (e.g., GoEmotions' \emph{annoyance} $\to$ \emph{anger}).
    \item \textbf{Intersection}: retain only labels with exact string matches in both taxonomies, dropping rows whose gold labels fall entirely outside this set.
\end{itemize}
Full label mappings are provided in \Cref{app:mappings}.

\section{Results}\label{sec:results}

We evaluate all six models on our in-domain test set and on three external benchmarks. For cross-benchmark evaluation, we report results in the intersection label space (exact string match between taxonomies), which is the stricter and less subjective protocol. Full projected-space results are provided in \Cref{app:full}.

\subsection{In-Domain Performance}\label{sec:indomain}

\Cref{tab:indomain} presents results on the 11,500-row test set (500 per language $\times$ 23~languages). \modelname{XLM-R-Large} leads on every metric, achieving 0.868~F1-micro and 0.987~AUC-micro. The three 278M-parameter base models (\modelname{XLM-R-Base}, \modelname{Twitter-XLM-R}, \modelname{mDeBERTa-v3}) are effectively tied at $\approx$0.839~F1-micro, despite differences in pretraining data and architecture. At the other end of the spectrum, \modelname{DistilBERT} trades approximately 8~F1 points for 9$\times$ faster training than \modelname{XLM-R-Large}, making it a viable option for latency-constrained deployments.

\begin{figure}[t]
    \centering
    \includegraphics[width=\linewidth]{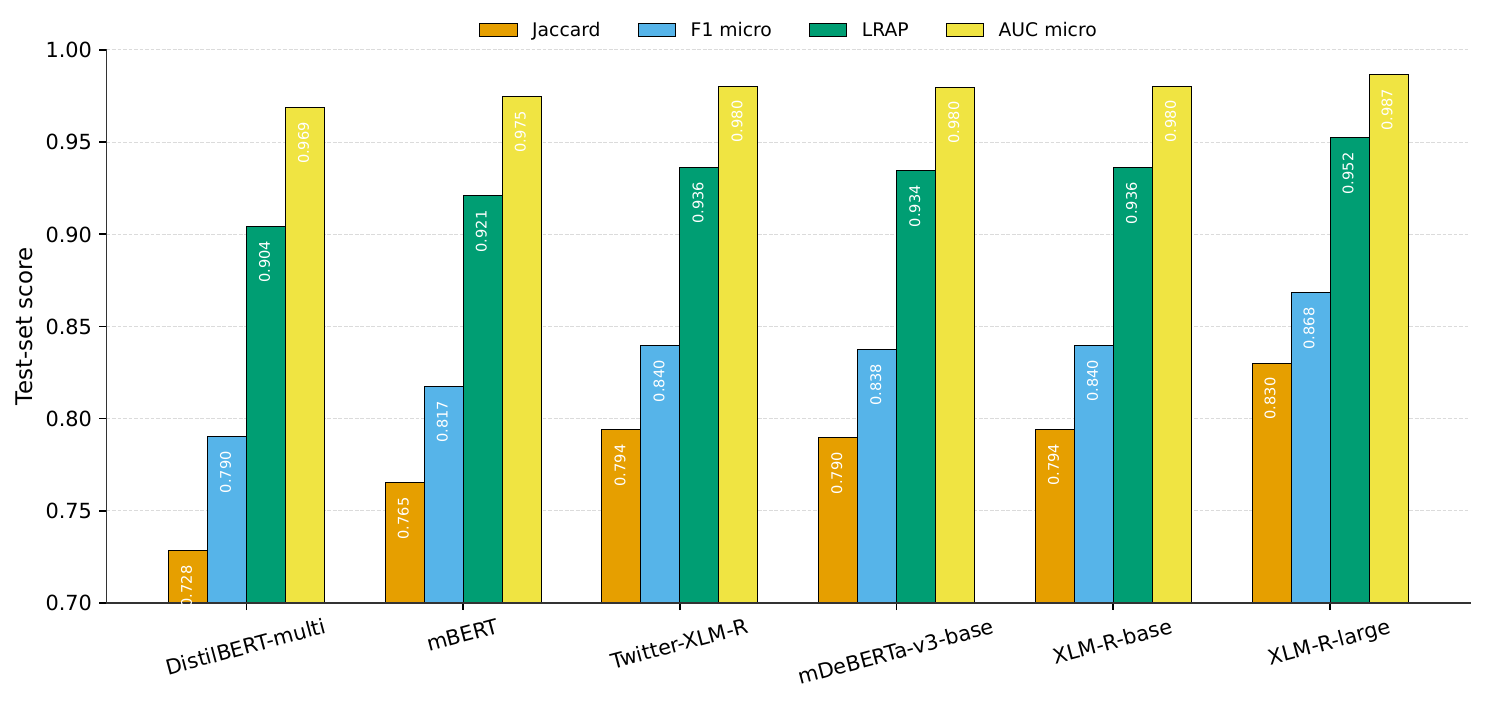}
    \caption{In-domain test performance across four key metrics. \modelname{XLM-R-Large} leads on all metrics; the three base models cluster tightly.}\label{fig:headline}
\end{figure}

\begin{figure}[t]
    \centering
    \includegraphics[width=\linewidth]{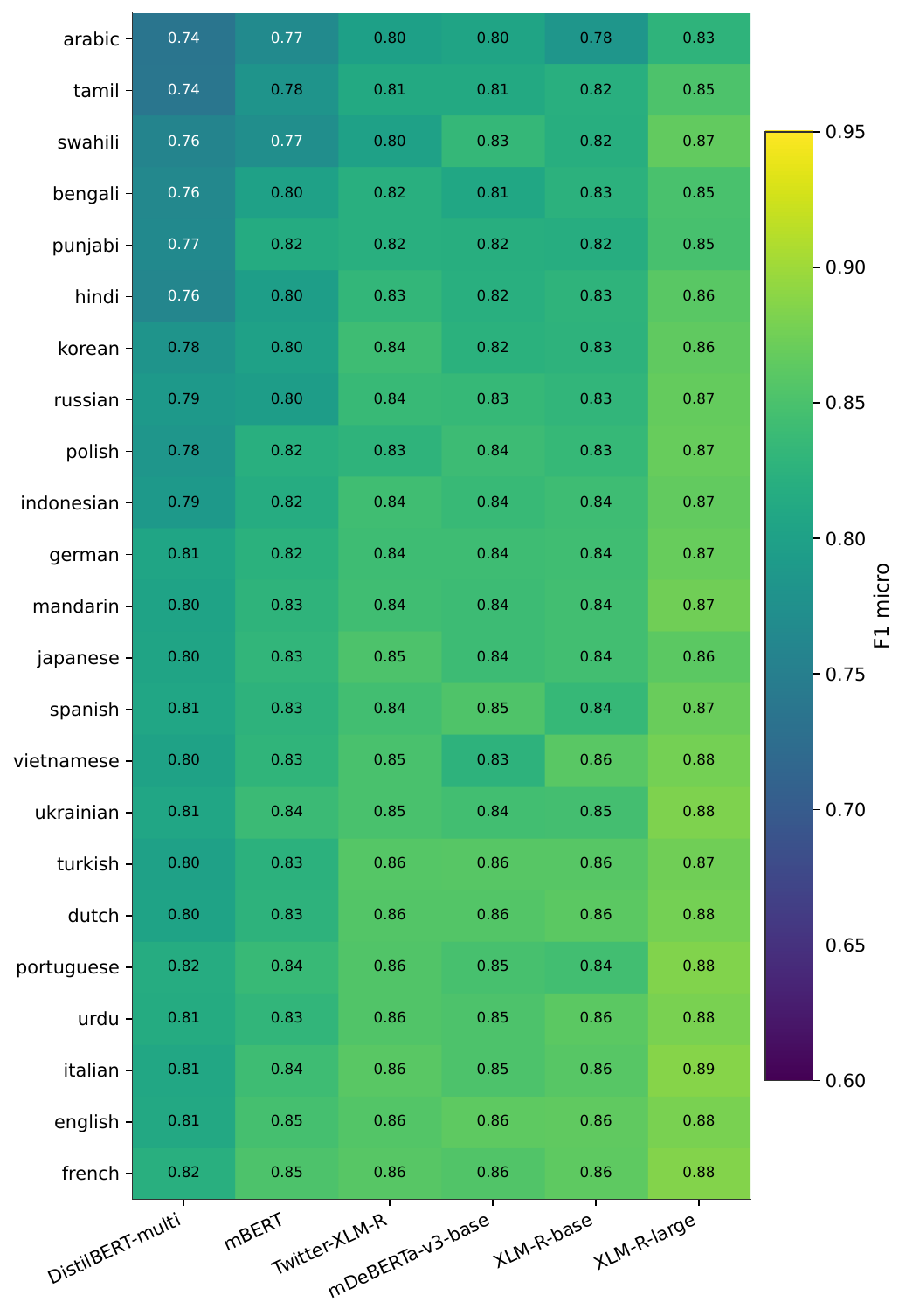}
    \caption{Per-language F1-micro on the test set. Languages are sorted by mean F1 across models (hardest at top).}\label{fig:perlang}
\end{figure}

\Cref{fig:headline} visualises the four most-reported metrics across all models. \Cref{fig:perlang} shows the per-language F1-micro heatmap; the model ranking is consistent across all 23~languages, though languages with non-Latin scripts exhibit slightly larger inter-model variance.

\begin{table}[t]
\centering
\caption{In-domain test results (23 languages, 11 emotions, $\tau = 0.5$). Best in \textbf{bold}.}\label{tab:indomain}
\small
\begin{tabular}{lcccccc}
\toprule
\textbf{Model} & \textbf{F1-mic} & \textbf{F1-mac} & \textbf{Jacc.} & \textbf{AUC} & \textbf{AP} & \textbf{LRAP} \\
\midrule
\modelname{XLM-R-Large}     & \metricbf{.868} & \metricbf{.868} & \metricbf{.830} & \metricbf{.987} & \metricbf{.946} & \metricbf{.952} \\
\modelname{XLM-R-Base}      & .840 & .839 & .794 & .980 & .923 & .936 \\
\modelname{Twitter-XLM-R}   & .840 & .839 & .794 & .980 & .922 & .936 \\
\modelname{mDeBERTa-v3}     & .838 & .837 & .790 & .980 & .920 & .934 \\
\modelname{mBERT}            & .817 & .816 & .765 & .975 & .903 & .921 \\
\modelname{DistilBERT}       & .790 & .789 & .728 & .969 & .881 & .904 \\
\bottomrule
\end{tabular}
\end{table}

\subsection{Cross-Benchmark Transfer}\label{sec:cross}

To assess whether models trained on synthetic data generalise to human-annotated corpora, we evaluate all models zero-shot on GoEmotions~\cite{demszky2020goemotions} (English) and SemEval-2018 E-c~\cite{mohammad2018semeval} (English, Arabic, Spanish). \Cref{tab:cross} reports results in the intersection label space.

On GoEmotions (9~shared labels, 3,144~rows), \modelname{XLM-R-Large} achieves 0.534~F1-micro and 0.877~AUC-micro. On SemEval English (7~shared labels), it reaches 0.436~F1-micro and 0.806~AUC-micro. In both cases, the model ranking from the in-domain evaluation is largely preserved.

An instructive finding emerges on the non-English SemEval subsets. On Arabic, \modelname{Twitter-XLM-R} narrowly outperforms \modelname{XLM-R-Large} (0.520 vs.\ 0.503~F1-micro), consistent with its base pretraining on tweet-like text. On Spanish, \modelname{mDeBERTa-v3} leads (0.467 vs.\ 0.466), though the margin is negligible. These results suggest that for specific language-domain pairs, the pretraining distribution of the base model can matter more than raw parameter count.

\begin{table}[t]
\centering
\caption{Cross-benchmark transfer (intersection label space). Best per benchmark in \textbf{bold}. GoE = GoEmotions, SE = SemEval-2018 E-c.}\label{tab:cross}
\small
\begin{tabular}{l cc cc cc cc}
\toprule
& \multicolumn{2}{c}{\textbf{GoE (EN)}} & \multicolumn{2}{c}{\textbf{SE (EN)}} & \multicolumn{2}{c}{\textbf{SE (AR)}} & \multicolumn{2}{c}{\textbf{SE (ES)}} \\
\cmidrule(lr){2-3}\cmidrule(lr){4-5}\cmidrule(lr){6-7}\cmidrule(lr){8-9}
\textbf{Model} & F1 & AUC & F1 & AUC & F1 & AUC & F1 & AUC \\
\midrule
\modelname{XLM-R-Large}     & \metricbf{.534} & \metricbf{.877} & \metricbf{.436} & \metricbf{.806} & .503 & \metricbf{.863} & .466 & \metricbf{.835} \\
\modelname{Tw-XLM-R}    & .518 & .859 & .401 & .789 & \metricbf{.520} & .860 & .457 & .824 \\
\modelname{XLM-R-B}     & .517 & .855 & .389 & .770 & .506 & .842 & .429 & .799 \\
\modelname{mDeBERTa}    & .494 & .846 & .409 & .784 & .503 & .837 & \metricbf{.467} & .821 \\
\modelname{mBERT}        & .458 & .819 & .371 & .741 & .392 & .785 & .373 & .747 \\
\modelname{DistilBERT}   & .440 & .809 & .315 & .728 & .361 & .786 & .334 & .738 \\
\bottomrule
\end{tabular}
\end{table}

\subsection{Head-to-Head with English Specialists}\label{sec:headtohead}

We compare our models against two widely-used English emotion classifiers evaluated on their native label subsets: \modelname{bhadresh-distilbert}\footnote{https://huggingface.co/bhadresh-savani/distilbert-base-uncased-emotion}~(6~labels: sadness, joy, love, anger, fear, surprise) and \modelname{j-hartmann-distilroberta}\footnote{https://huggingface.co/j-hartmann/emotion-english-distilroberta-base}~(7~labels: anger, disgust, fear, joy, neutral, sadness, surprise). Both are single-label softmax classifiers evaluated with argmax; our models use sigmoid $\geq 0.5$.

This difference in decision rule has an important consequence. On predominantly single-label benchmarks, argmax inflates F1 because it always predicts exactly one label, matching the gold cardinality by construction. Threshold-free ranking metrics (AUC, AP, LRAP) are immune to this artefact and provide the fairer comparison.

\Cref{tab:headtohead} shows the comparison on SemEval~EN, which is the benchmark least likely to overlap with either model's training data. On bhadresh's 6-label subset, \modelname{XLM-R-Large} trails on F1-micro (0.467 vs.\ 0.557) but \emph{matches} on AP-micro (0.636) and LRAP (0.804), and \emph{surpasses} on AUC-micro (0.810 vs.\ 0.787). On j-hartmann's 7-label subset, \modelname{XLM-R-Large} again achieves the highest AUC-micro (0.742 vs.\ 0.700). We find that our synthetically-trained multilingual models rank emotions as accurately as monolingual English specialists, while additionally supporting 22~other languages where those specialists cannot operate at all.

\begin{table}[t]
\centering
\caption{Head-to-head on SemEval~EN test in each specialist's label subset. $\dag$ = English-only single-label specialist. Highest per metric in \textbf{bold}.}\label{tab:headtohead}
\small
\begin{tabular}{l cccc}
\toprule
\textbf{Model} & \textbf{F1-mic} & \textbf{AUC-mic} & \textbf{AP-mic} & \textbf{LRAP} \\
\midrule
\multicolumn{5}{l}{\emph{bhadresh's 6-label subset}} \\
\modelname{bhadresh}$^\dag$     & \metricbf{.557} & .787 & \metricbf{.636} & \metricbf{.804} \\
\modelname{XLM-R-Large} (ours) & .467 & \metricbf{.810} & \metricbf{.636} & \metricbf{.804} \\
\modelname{mDeBERTa-v3} (ours) & .438 & .792 & .616 & .793 \\
\midrule
\multicolumn{5}{l}{\emph{j-hartmann's 7-label subset}} \\
\modelname{j-hartmann}$^\dag$   & \metricbf{.461} & .700 & \metricbf{.518} & \metricbf{.712} \\
\modelname{XLM-R-Large} (ours) & .375 & \metricbf{.742} & .473 & .705 \\
\modelname{mDeBERTa-v3} (ours) & .356 & .726 & .466 & .699 \\
\bottomrule
\end{tabular}
\end{table}

\subsection{Compute-Quality Trade-off}

\begin{figure}[t]
    \centering
    \includegraphics[width=0.8\linewidth]{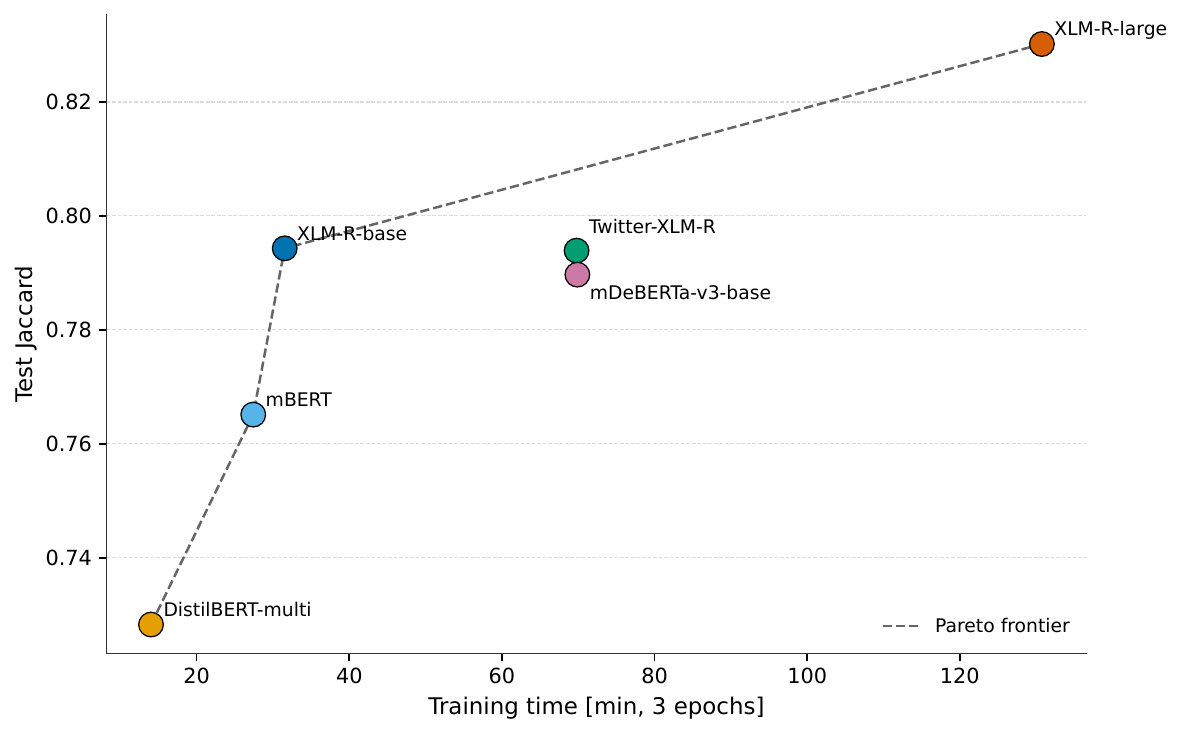}
    \caption{Compute vs.\ quality Pareto frontier. Training time (min, log scale) vs.\ test Jaccard. \modelname{XLM-R-Base} dominates among base models due to its 2$\times$ speed advantage.}\label{fig:pareto}
\end{figure}

\Cref{fig:pareto} shows the Pareto frontier of training cost versus in-domain Jaccard. \modelname{DistilBERT} occupies the efficiency corner (14~min, 0.728~Jaccard), while \modelname{XLM-R-Large} anchors the quality frontier (131~min, 0.830~Jaccard). Among the base models, \modelname{XLM-R-Base} offers the best cost-quality ratio: it trains in 31.5~min, roughly half the time of \modelname{mDeBERTa-v3} and \modelname{Twitter-XLM-R}, while matching their accuracy. This makes it the natural choice for practitioners who need a strong multilingual emotion classifier without the computational overhead of a large model.

\begin{figure}[t]
    \centering
    \includegraphics[width=0.8\linewidth]{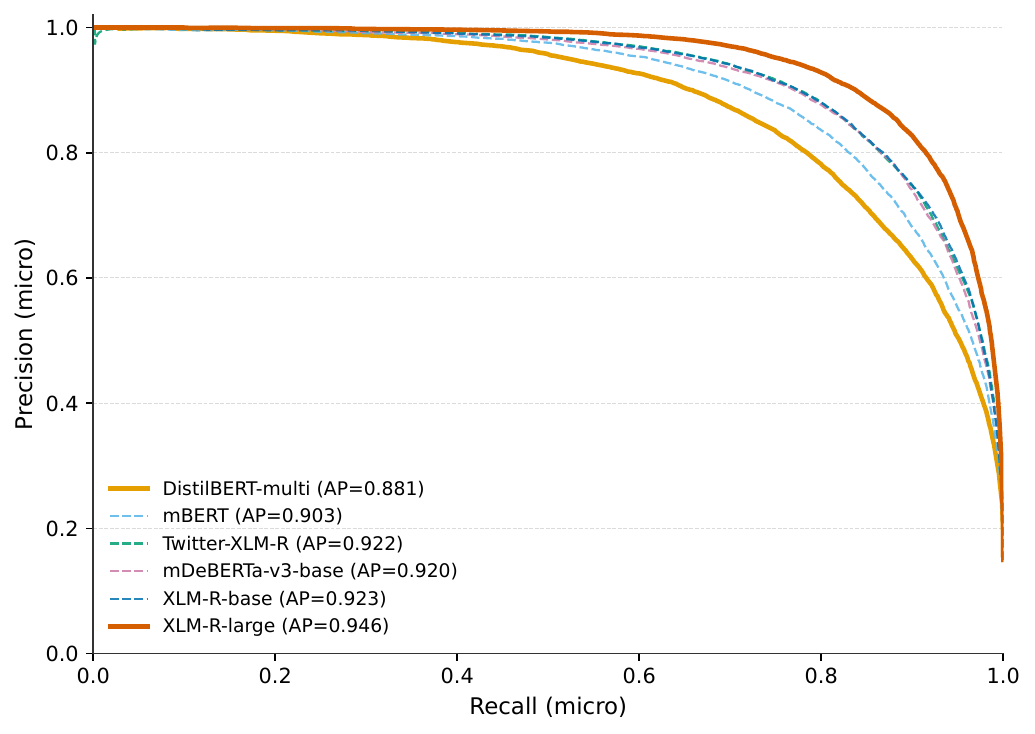}
    \caption{Micro-averaged precision-recall curves on the in-domain test set. AP-micro values are shown in the legend.}\label{fig:pr}
\end{figure}

\section{Discussion}\label{sec:discussion}

\paragraph{Synthetic data enables competitive multilingual models.}
We find that models trained entirely on synthetic data achieve ranking-metric parity with English specialists on real benchmarks, while covering 23~languages. The F1 gap observed in cross-benchmark evaluation is largely attributable to the sigmoid vs.\ argmax asymmetry rather than inferior emotion understanding: our models rank emotions correctly but predict multi-label sets where the gold standard is predominantly single-label.

\paragraph{Pretraining distribution matters.}
The Arabic SemEval result, where \modelname{Twitter-XLM-R} outperforms the larger \modelname{XLM-R-Large}, illustrates that domain match in base pretraining can compensate for model capacity. For deployment scenarios involving social media text, domain-adapted base models may be preferable to larger general-purpose ones.

\paragraph{Limitations.}
We identify three limitations of the current approach. First, synthetic data inevitably reflects the biases and stylistic patterns of the generation process. In particular, LLMs tend to produce longer and more verbose outputs than naturally-occurring text \cite{borisov2026chatbot}, which may shift the length and style distribution of the training corpus relative to real-world inputs such as tweets or short reviews. Second, our label taxonomy omits emotions like \emph{anticipation} and fine-grained states like \emph{embarrassment} and \emph{pride}, which must be collapsed into coarser categories during cross-benchmark evaluation. Third, we do not conduct formal human evaluation of the synthetic data quality; while programmatic filtering and auxiliary-classifier checks provide quality assurance, annotator agreement studies would strengthen confidence in label fidelity.

\section{Related Work}\label{sec:related}

\paragraph{Emotion classification.}
Early work on emotion classification focused on Ekman's six basic emotions~\cite{ekman1992basic} using single-label annotation. GoEmotions~\cite{demszky2020goemotions} expanded the label set to 28~fine-grained categories with multi-label annotation but remains English-only. SemEval-2018 Task~1 E-c~\cite{mohammad2018semeval} introduced multi-label emotion classification across English, Arabic, and Spanish with 11~categories. Our work extends the multilingual scope to 23~languages while maintaining multi-label annotation and a taxonomy designed for cross-lingual applicability.

\paragraph{Synthetic data for NLP.}
The use of large language models for the generation of synthetic data (sometimes artificial data \cite{borisov2024open}) has shown promise in a range of NLP tasks, including instruction tuning~\cite{wang2023selfinstruct}, data enhancement~\cite{dai2023auggpt}, and low-resource language processing. Cultural adaptation of synthetic generation remains less explored \cite{gyamfi2026synthetic}. We contribute a methodology for producing culturally-grounded emotional text across diverse linguistic contexts, and demonstrate that the resulting data is sufficient to train competitive classifiers.

\paragraph{Multilingual transformers.}
The encoder models evaluated in this work, including mBERT~\cite{devlin2019bert}, XLM-RoBERTa~\cite{conneau2020xlmr}, and mDeBERTa~\cite{he2021deberta}, represent the standard multilingual encoder family. Our comparative study complements existing cross-lingual benchmarks such as XTREME~\cite{hu2020xtreme} by providing an emotion-specific evaluation across 23~languages under controlled training conditions.

\section{Conclusion}\label{sec:conclusion}

We have presented a pipeline for generating culturally-adapted synthetic emotion data across 23~languages and demonstrated that models trained on this data are competitive with English-only specialized models on real benchmarks. Our best model, \modelname{XLM-R-Large}, achieves 0.868~F1-micro in-domain and matches specialist models on threshold-free ranking metrics on GoEmotions and SemEval-2018 E-c, while being the only model that natively supports all 23~training languages. Among the base models, \modelname{XLM-R-Base} offers the strongest cost-quality trade-off, training in 31.5~minutes while matching models that require twice the compute. 

\bibliographystyle{unsrt}
\bibliography{refs}

\newpage
\appendix

\section{Label Mappings}\label{app:mappings}

\subsection{GoEmotions $\to$ Our Taxonomy (28 $\to$ 11)}

\begin{table}[h]
\centering
\small
\begin{tabular}{ll}
\toprule
\textbf{Our label} & \textbf{GoEmotions labels mapped} \\
\midrule
anger       & anger, annoyance \\
contempt    & disapproval \\
disgust     & disgust \\
fear        & fear, nervousness, embarrassment \\
frustration & disappointment \\
gratitude   & gratitude \\
joy         & joy, amusement, excitement, optimism, pride, relief, admiration, approval \\
love        & love, caring, desire \\
neutral     & neutral \\
sadness     & sadness, grief, remorse \\
surprise    & surprise, realization, curiosity, confusion \\
\bottomrule
\end{tabular}
\caption{GoEmotions label projection.}\label{tab:goe_mapping}
\end{table}

\subsection{SemEval-2018 E-c $\to$ Our Taxonomy (11 $\to$ 7)}

\begin{table}[h]
\centering
\small
\begin{tabular}{ll}
\toprule
\textbf{Our label} & \textbf{SemEval labels mapped} \\
\midrule
anger   & anger \\
disgust & disgust \\
fear    & fear \\
joy     & joy, optimism \\
love    & love, trust \\
sadness & sadness, pessimism \\
surprise & surprise \\
\bottomrule
\end{tabular}
\caption{SemEval-2018 E-c label projection. \emph{anticipation} is dropped (no cognate in our taxonomy).}\label{tab:semeval_mapping}
\end{table}

\section{Full Cross-Benchmark Results}\label{app:full}

\subsection{GoEmotions --- Projected (11-label space, 5,427 rows)}

\begin{table}[h]
\centering
\small
\begin{tabular}{lcccccc}
\toprule
\textbf{Model} & \textbf{F1-mic} & \textbf{F1-mac} & \textbf{Jacc.} & \textbf{AUC} & \textbf{AP} & \textbf{LRAP} \\
\midrule
\modelname{XLM-R-Large}     & .357 & .314 & .312 & .750 & .326 & .542 \\
\modelname{Twitter-XLM-R}   & .349 & .314 & .293 & .748 & .331 & .531 \\
\modelname{XLM-R-Base}      & .345 & .305 & .296 & .740 & .332 & .523 \\
\modelname{mDeBERTa-v3}     & .337 & .308 & .278 & .740 & .330 & .526 \\
\modelname{mBERT}            & .317 & .285 & .266 & .709 & .304 & .489 \\
\modelname{DistilBERT}       & .306 & .272 & .249 & .704 & .300 & .483 \\
\bottomrule
\end{tabular}
\caption{GoEmotions test---projected label space.}\label{tab:goe_proj}
\end{table}

\subsection{SemEval-2018 E-c --- Full Results by Language}

\begin{table}[h]
\centering
\small
\begin{tabular}{l cc cc cc}
\toprule
& \multicolumn{2}{c}{\textbf{English}} & \multicolumn{2}{c}{\textbf{Arabic}} & \multicolumn{2}{c}{\textbf{Spanish}} \\
\cmidrule(lr){2-3}\cmidrule(lr){4-5}\cmidrule(lr){6-7}
\textbf{Model} & F1-mic & AUC & F1-mic & AUC & F1-mic & AUC \\
\midrule
\modelname{XLM-R-Large}     & .366 & .756 & .424 & .804 & .366 & .782 \\
\modelname{Twitter-XLM-R}   & .342 & .743 & .446 & .806 & .366 & .777 \\
\modelname{XLM-R-Base}      & .331 & .727 & .441 & .798 & .344 & .758 \\
\modelname{mDeBERTa-v3}     & .348 & .739 & .435 & .792 & .377 & .776 \\
\modelname{mBERT}            & .318 & .705 & .329 & .734 & .304 & .719 \\
\modelname{DistilBERT}       & .271 & .682 & .307 & .725 & .272 & .703 \\
\bottomrule
\end{tabular}
\caption{SemEval-2018 E-c test---projected (11-label space), by language.}\label{tab:semeval_proj}
\end{table}

\end{document}